\documentclass[letterpaper]{article}

\usepackage[preprint]{aaai2027}

\usepackage[hyphens]{url}
\usepackage{graphicx}
\usepackage{natbib}
\usepackage{caption}
\usepackage{amsmath}
\usepackage{amssymb}
\usepackage{algorithm}
\usepackage{algorithmic}
\usepackage{booktabs}

\title{Codec-Gauge: Learning Compression-Friendly Gauges for Transformer KV Caches}

\author{
  Yitao Jiang\textsuperscript{\rm 1},
  Yaoqing Yang\textsuperscript{\rm 1},
  Luyang Zhao\textsuperscript{\rm 2},
  Muhao Chen\textsuperscript{\rm 3},
  Devin Balkcom\textsuperscript{\rm 1}
}
\affiliations{
  \textsuperscript{\rm 1}Department of Computer Science, Dartmouth College, Hanover, NH\\
  \textsuperscript{\rm 2}Department of Electrical and Computer Engineering, Clemson University, Clemson, SC\\
  \textsuperscript{\rm 3}Department of Mechanical and Aerospace Engineering, University of Houston, Houston, TX\\
  \textsuperscript{\rm 1}\{yitao.jiang.gr, yaoqing.yang, devin.balkcom\}@dartmouth.edu\\
  \textsuperscript{\rm 2}luyangz@clemson.edu\\
  \textsuperscript{\rm 3}muhaochen@uh.edu
}

\begin{document}

\maketitle

\begin{abstract}
Long-context Transformer inference increasingly relies on KV-cache compression or quantization. Rotation and transform-coding results indicate that the channel basis of each key/value vector can affect how faithfully a fixed backend preserves model behavior. We introduce \emph{Codec-Gauge}, a post-training cache-coordinate layer that learns small orthogonal channel transforms, or gauges, around existing backends. Its frequency-distribution objective combines a token--channel DCT spectral-centroid loss with a smooth rate proxy to concentrate KV energy in low-frequency codec-facing layouts. We evaluate actual compression/decompression with measured bytes and rolling compressed-history scoring. Across six models at 3, 4, and 6 bits/value, learned gauges reduce zfp KL divergence by $44.0\%$ on average relative to raw coordinates and outperform random, Hadamard, DCT, and PCA/KLT controls; the same gauges improve quality preservation for block-uniform and KIVI-style quantization. A 27B extension and task-prompt likelihoods reproduce the quality trend, while serial storage/timing measurements validate the implemented compressed-cache paths.
\end{abstract}

\section{Introduction}

\begin{figure}[t]
\centering
\includegraphics[width=\columnwidth]{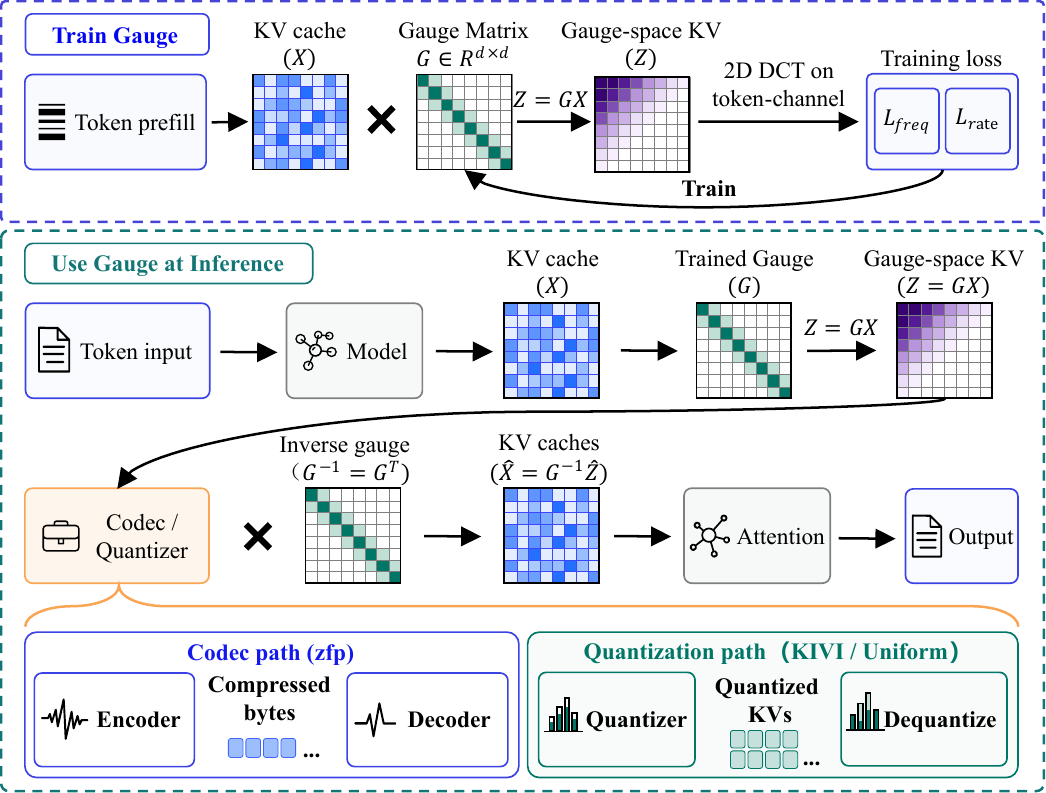}
\caption{Overview of Codec-Gauge. A one-time post-training stage learns orthogonal gauges from frozen-model KV tensors. At inference, the gauge maps KV tensors into codec-facing coordinates before compression or quantization, and the inverse gauge restores recovered KV states before attention.}
\label{fig:overview}
\end{figure}

Long-context Transformer inference is increasingly constrained by the memory and bandwidth cost of KV caches. During autoregressive decoding, each layer stores past key and value tensors, and this cache grows linearly with batch size, context length, number of layers, number of KV heads, and head dimension. Attention kernels and memory managers improve access locality, while multi-query and grouped-query attention reduce stored KV heads (\citep{shazeer2019mqa,ainslie2023gqa,dao2022flashattention,kwon2023vllm}); nevertheless, the cache remains a large runtime object that must be stored, moved, and repeatedly consumed. This paper isolates a complementary variable: the coordinate basis presented to a compression or quantization backend. Codec-Gauge learns this basis from frozen-model KV tensors to improve fidelity at the same measured rate while leaving model weights, attention semantics, and backend coding rules unchanged (Figure~\ref{fig:overview}).

A large body of work reduces this cost by changing cache precision, cache retention, or model representation (\citep{liu2024kivi,hooper2024kvquant,kang2024gear,yang2024mikv,lin2025qserve,zhang2023h2o,xiao2024streamingllm,li2024snapkv,deepseek2024v2,chang2025palu,nawrot2024dmc,gelberg2026kvcat}). Codec-Gauge is motivated by two related observations: transform coding shows that a basis change can concentrate signal energy into coefficients that are easier to preserve at a fixed rate (\citep{ahmed1974dct,lindstrom2014zfp}), and LLM quantization shows that orthogonal channel rotations can change low-bit error behavior without changing the represented function (\citep{chee2023quip,tseng2024quipsharp,ashkboos2024quarot,liu2025spinquant,su2025rotatekv,saxena2025kvlinc}). Together, these observations motivate learning, after pretraining, the representation geometry presented to a fixed backend. Although model channels have no native spatial order, a codec consumes a concrete memory layout, so the learned channel basis can change the smoothness and coefficient structure of the stored KV field.

This paper studies geometry as a first-class compression variable. \emph{Codec-Gauge} learns an orthogonal gauge over KV channel groups so that existing compression and quantization backends see a more compression-friendly representation, while model weights, attention semantics, and backend implementations remain fixed. By KV coordinates, we mean the channel basis used to represent each key or value vector within a head. The gauge is an invertible change of this basis: it can mix channel coordinates within each group, but it never mixes tokens, heads, or layers, and it never removes cache entries. At inference time, cached keys and values are mapped into the learned gauge before compression or quantization and mapped back before attention. Along a no-error path, the gauge and inverse preserve the cache up to numerical precision; under lossy backends, the gauge changes the geometry seen by the backend and therefore changes recovery error and quality retained after decompression.

We use a matched-rate experimental design. Within each comparison, the model, token stream, backend, measured bit budget, and evaluator are fixed, and only the cache-coordinate map is varied. This makes identity, random orthogonal, fixed transform, data-derived, and learned gauges comparable under the same backend, so improvements can be attributed to cache geometry rather than to a changed quantizer, eviction rule, packing format, kernel, or scheduler.

Our main target is fixed-rate numerical compression of continuous KV tensors. We use GPU zfp compression as the primary codec because its block-transform structure makes coefficient concentration and local smoothness directly relevant to reconstruction quality at a fixed rate (\citep{lindstrom2014zfp}). The key technical choice is a frequency-distribution loss on the codec-facing token--channel layout: a DCT spectral-centroid term moves KV energy toward lower frequencies, while a smooth log-amplitude proxy favors coefficient distributions with fewer significant transform coefficients. All quality claims use actual compression/decompression and rolling compressed-history scoring, where newly created KV entries traverse the same gauge and backend path.

The codec remains outside the training loop: fitting the gauge uses no reconstruction, language-modeling, logit, or task-output loss.

We use coordinate controls to identify the effect of this training objective. Identity uses the raw cache basis, while random orthogonal gauges capture generic rotation and outlier-redistribution effects known from quantization (\citep{chee2023quip,tseng2024quipsharp,ashkboos2024quarot,liu2025spinquant,sun2025flatquant,su2025rotatekv,saxena2025kvlinc}). Hadamard and DCT provide fixed structured transforms, PCA/KLT provides a data-derived basis, and learned gauges test the proposed frequency-distribution objective under the same backend and measured rates. These controls separate codec-facing gauge learning from both raw-coordinate compression and generic orthogonal preprocessing.

\paragraph{Backend compatibility.}
The learned gauge is a fixed per-checkpoint transform, not a per-token payload, and changes neither a backend's coding rule nor its bit budget. It is applied at the KV tensor interface; our measurements use explicit recovery and include that cost in timing, tying quality, storage, and latency to the same implemented path.

Across six language models and multiple matched bits-per-value settings, learned gauges reduce KV reconstruction error and rolling suffix degradation for zfp beyond identity, random, fixed-transform, and PCA/KLT controls. In the same trained coordinates, block-uniform and KIVI-style quantizers (\citep{liu2024kivi}) also preserve output quality better than their raw-coordinate versions. A larger 27B model check and task-prompt likelihoods reproduce the coordinate effect, while serial context-length measurements verify realized storage and recovery cost. For zfp, learned gauges reduce KL divergence, logit MSE, top-1 flip rate, and KV NRMSE by $44.0\%$, $43.3\%$, $24.5\%$, and $18.3\%$ at the main operating points.

This paper makes four contributions. First, we formulate KV-cache coordinate geometry as a post-training variable for improving compression fidelity under fixed backends. Second, we introduce Codec-Gauge, an orthogonal cache-coordinate layer that wraps existing compression and quantization backends without changing attention semantics. Third, we design a frequency-distribution objective that trains this layer from frozen KV tensors using codec-facing token--channel spectral structure, without language-modeling loss, logit matching, task supervision, or model-weight updates. Fourth, we provide paired evidence that learned gauges improve actual zfp GPU compression under matched measured rates and also improve quality retention for block-uniform and KIVI-style quantization paths.

\section{Related Work}

\paragraph{KV-cache quantization and rotations.}
KV-cache quantization reduces stored key/value precision through asymmetric granularity, outlier handling, correction terms, mixed precision, or serving co-design (\citep{liu2024kivi,hooper2024kvquant,kang2024gear,yang2024mikv,lin2025qserve}). Coordinate transforms are also central to LLM quantization, including incoherence processing, outlier-removing rotations, learned rotations, distribution flattening, adaptive KV rotations, and Hadamard-based correction (\citep{chee2023quip,tseng2024quipsharp,ashkboos2024quarot,liu2025spinquant,sun2025flatquant,su2025rotatekv,saxena2025kvlinc}). Codec-Gauge uses a cache-geometric frequency objective for a fixed numerical codec rather than a scalar quantization loss, then reuses the same trained coordinates for scalar low-bit paths.

\paragraph{Token, page, and dynamic cache selection.}
Another line of work reduces the number of cache entries retained or accessed, using heavy hitters, attention sinks, model-internal patterns, prompt observations, layer-wise budgets, or query-aware sparsity (\citep{zhang2023h2o,xiao2024streamingllm,ge2023fastgen,li2024snapkv,cai2025pyramidkv,tang2024quest}). InfiniGen, RocketKV, and CacheGen emphasize dynamic cache management, multi-stage compression, or cache streaming (\citep{lee2024infinigen,behnam2025rocketkv,liu2024cachegen}). Codec-Gauge instead operates on the dense tensor coordinates of retained entries, so it can be layered with retention and paging policies.

\paragraph{Transform and numerical KV compression.}
Frequency-domain and transform-coding methods treat KV caches as structured numerical signals. FreqKV, FAEDKV, and KVTC exploit frequency or transform-domain redundancy (\citep{kai2026freqkv,li2025faedkv,staniszewski2026kvtc}), while DCT and zfp show how spectral concentration and local smoothness affect array rate-distortion behavior (\citep{ahmed1974dct,lindstrom2014zfp}). Codec-Gauge contributes a learned coordinate layer for this setting: the gauge shapes the numerical field seen by an existing codec's token--channel layout, and compression is performed by the backend itself.

\paragraph{Architectural, low-rank, and training-aware KV representations.}
Several methods reduce KV state by changing model structure or learning a different cache representation, including latent KV states, low-rank projection, depth redundancy, dynamic memory compression, training for compressible KV, and adaptive orthogonal projections (\citep{deepseek2024v2,chang2025palu,liu2024minicache,nawrot2024dmc,gelberg2026kvcat,lin2025matryoshkakv}). Codec-Gauge keeps the checkpoint frozen: a small invertible gauge is trained for an existing model and evaluated around an explicit compression/decompression path.

\section{Problem Statement and Overview}

\begin{table}[t]
\centering
\small
\begin{tabular}{@{}lp{0.68\columnwidth}@{}}
\toprule
Symbol & Meaning \\
\midrule
$B$ & Batch size \\
$T$ & Prefix/context length stored in cache \\
$L$ & Number of standard softmax-attention KV layers \\
$H_l$ & Number of KV heads in layer $l$ \\
$d_l$ & Head dimension in layer $l$ \\
$K_l,V_l$ & Key and value caches in layer $l$ \\
$X_{l,s}$ & Unified cache tensor, $s\in\{K,V\}$ \\
$g$ & Gauge channel group dimension; $g=d_l$ is full-head \\
$R_l$ & Number of channel groups in layer $l$ \\
$G_{l,s,h,r}$ & Gauge for layer/type/head/group \\
$Z_{l,s}$ & Gauge-space cache tensor \\
$C,D$ & Codec compression and decompression maps \\
$B_c$ & Compressed byte buffer \\
$\hat{Z}_{l,s},\hat{X}_{l,s}$ & Decoded transformed-coordinate and recovered cache tensors \\
\bottomrule
\end{tabular}
\caption{Notation for cache gauge training.}
\label{tab:notation}
\end{table}

Table~\ref{tab:notation} summarizes the notation. We consider the standard key/value cache produced by softmax-attention Transformers during autoregressive inference. For layer $l$, the cache for the previous $T$ tokens is:
\[
K_l,V_l \in \mathbb{R}^{B\times H_l\times T\times d_l}, \qquad
\mathcal{M}_T=\{(K_l,V_l)\}_{l=1}^{L}.
\]
The number of cache elements in FP16 or BF16 storage is:
\[
N_{\mathrm{KV}}(T)=\sum_{l=1}^{L}2BH_lTd_l,
\]
which grows linearly with context length. We only consider key/value tensors exposed by standard softmax attention; recurrent or linear-attention states are outside the gauge and compression object studied here.

Codec-Gauge learns a cache-coordinate gauge without changing tokens, attention architecture, or model weights. Let $X_{l,s}$ denote a layer-$l$ cache tensor with $s\in\{K,V\}$. For each layer, cache type, head, and channel group, we learn an invertible gauge $G_{l,s,h,r}$. The group dimension $g$ partitions the head dimension into $R_l=d_l/g$ groups; when $g=d_l$, the gauge acts on the full head. For group $r=1,\ldots,R_l$,
\[
X_{l,s}[b,h,t,r]\in\mathbb{R}^{g},
\]
and the transformed cache is:
\[
Z_{l,s}[b,h,t,r]=G_{l,s,h,r}X_{l,s}[b,h,t,r].
\]
The transform mixes channels only within a group; it neither mixes nor removes tokens.

At inference time, the backend observes the transformed-coordinate cache rather than the raw cache:
\[
\begin{aligned}
Z &= GX, & B_c &= C(Z),\\
\hat{Z} &= D(B_c), & \hat{X} &= G^{-1}\hat{Z},
\end{aligned}
\]
where $C$ and $D$ are the compression/decompression or quantization/reconstruction maps of a fixed backend, and $B_c$ is the compressed buffer consumed only through the decoded cache $\hat{X}$. Without backend error, $G^{-1}GX=X$ and model behavior is unchanged up to numerical precision. Under lossy backends, the gauge leaves attention semantics fixed but changes the numerical geometry seen by the backend, and therefore the structure of recovery error and downstream output perturbation, meaning changes in likelihoods, logits, and top-token rankings relative to the full-cache reference.

The evaluation target is direct: for a frozen Transformer, fixed data distribution, fixed backend, and matched actual bits/value, a learned gauge should reduce KV reconstruction error and rolling suffix output perturbation relative to raw and control coordinates.

\section{Method and Evaluation Design}

\subsection{Gauge Parameterization and Training}

Given a frozen Transformer $f_\theta$, Codec-Gauge learns only cache-coordinate parameters and does not update model weights. For each layer $l$, cache type $s\in\{K,V\}$, head $h$, and channel group $r$, we parameterize an orthogonal gauge as:
\[
G_{l,s,h,r}=\exp(A_{l,s,h,r}-A_{l,s,h,r}^{\top}),\qquad
G_{l,s,h,r}^{-1}=G_{l,s,h,r}^{\top},
\]
where $\exp(\cdot)$ is the matrix exponential. Since $A-A^\top$ is skew-symmetric, the resulting small group-level gauge is orthogonal and invertible without an orthogonality penalty. The group dimension $g$ sets the channel subspace covered by each gauge: smaller groups restrict mixing, while full-head gauges provide higher expressivity. We evaluate $g=8$, $16$, $32$, and full-head gauges, and report $g=16$ as the representative grouped setting.

The gauge is a fixed model-side object, not a per-token cache payload. Explicit application costs $O(N_{\mathrm{KV}}g)$ multiply-adds, and learned parameters scale as $O(\sum_l 2H_ld_lg)$ rather than with context length. Some recovery operations can be folded into downstream projections; key-side fusion depends on positional encoding and kernel layout. Our implementation uses explicit key/value recovery and charges that cost in timing.

Gauge training uses FineWeb-Edu sample-10BT (\citep{penedo2024fineweb}). For each model, we collect $64$ non-overlapping 4096-token windows, run frozen-model prefill, collect standard softmax-attention KV tensors, and optimize the gauge for $25$ epochs with AdamW. This one-time per-checkpoint step uses $T_b=16$, $\lambda_f=1.0$, $\lambda_r=0.02$, and $\tau=0.04$ for every checkpoint, with no model-specific tuning. Evaluation windows are recorded from disjoint FineWeb-Edu documents with a different seed. The objective depends only on cache geometry; it does not include language-modeling loss, logit matching, or task-output supervision.

\subsection{Gauge Training Objective}

For a transformed cache tensor $X'=GX$, we partition the token dimension into blocks of length $T_b$ and apply a two-dimensional orthonormal DCT to each $T_b\times d_l$ token--channel block:
\[
U=\mathrm{DCT}_{2D}(\mathrm{block}(X')).
\]
The training objective is
\[
L_{\mathrm{train}}
=
\lambda_f L_{\mathrm{freq}}
+\lambda_r L_{\mathrm{rate}} .
\]

The frequency term is a spectral centroid under the token--channel layout exposed to the codec. Although the gauge may be grouped, this statistic spans the complete head layout presented to zfp. The DCT supplies a differentiable statistic, and the gauge chooses a channel basis whose observed KV energy becomes concentrated in low token and layout frequencies. Channel frequency refers to memory order after the basis change, not intrinsic spatial order; token order remains unchanged. For coefficient position $(u,v)$, define the normalized radius
\[
\rho(u,v)=
\frac{1}{\sqrt{2}}
\sqrt{
\left(\frac{u}{T_b-1}\right)^2+
\left(\frac{v}{d_l-1}\right)^2 } .
\]
With coefficient energy $E_{u,v}=U_{u,v}^2$, we define
\[
L_{\mathrm{freq}}
=
\frac{\sum_{u,v}E_{u,v}\rho(u,v)}
{\sum_{u,v}E_{u,v}+\epsilon}.
\]
Minimizing this term shifts energy toward lower radial frequencies in the codec-facing layout. Since the gauge acts only on channel groups, it learns how each token-frequency component is represented across channel-layout modes, making neighboring coordinates in the fixed codec layout carry more correlated, low-frequency energy. Token order, channel grouping, and the zfp field layout are fixed across identity, random, fixed-transform, PCA/KLT, and learned conditions, so the coordinate map is the isolated variable. We track spectral concentration as $1-L_{\mathrm{freq}}$.

The rate term is a smooth log-amplitude proxy:
\[
L_{\mathrm{rate}}
=
\mathbb{E}_{u,v}
\left[
\log\left(1+\frac{|U_{u,v}|}{\tau}\right)
\right],
\]
with $\tau=0.04$. Because the gauge and DCT are energy-preserving transforms, this concave penalty favors concentrating energy into fewer significant coefficients. Together with $L_{\mathrm{freq}}$, it provides a differentiable codec-layout objective; held-out evaluation uses measured backend bytes and post-decode model behavior rather than the proxy itself. After training, the gauge is fixed for all evaluation windows.

\subsection{Compression Conditions}

The main evaluation grid evaluates each backend under identity, random, and learned coordinates. Identity compresses raw KV cache; random uses a sampled orthogonal gauge with the same block structure as the learned gauge; learned uses the trained gauge. Random captures the known benefits of generic rotation and outlier redistribution in quantization (\citep{chee2023quip,ashkboos2024quarot,liu2025spinquant,su2025rotatekv,saxena2025kvlinc}), so gains beyond it test the value of the learned codec-layout geometry.

The primary backend is GPU zfp compression applied to continuous KV tensors at 2-, 3-, 4-, 6-, and 8-bit/value. The implementation encodes contiguous FP32-transformed tensors with native CUDA zfp, decodes them, applies the inverse gauge, and records the actual compressed bytes. The CUDA extension presents each tensor as a two-dimensional zfp field with $n_x=d_l$ and $n_y=\mathrm{numel}/d_l$; reported bytes include the zfp payload and a fixed shape header. Block-uniform uses one range per $T_b\times d_l$ block, whereas the KIVI-style path uses per-channel key and per-token value ranges (\citep{liu2024kivi}); both use the same bit settings. Each main evaluation batch contains $49$ conditions: one full-cache baseline, three no-compression sanity rows, and $15$ rows each for zfp, block-uniform quantization, and KIVI-style quantization. A separate zfp coordinate-control run evaluates identity, learned, three random gauges, Hadamard, DCT, and PCA/KLT controls at $3$, $4$, and $6$ bits/value. PCA/KLT uses an uncentered second-moment eigenbasis computed at the same layer, K/V, head, and group granularity.

\begin{figure*}[t]
\centering
\includegraphics[width=\textwidth]{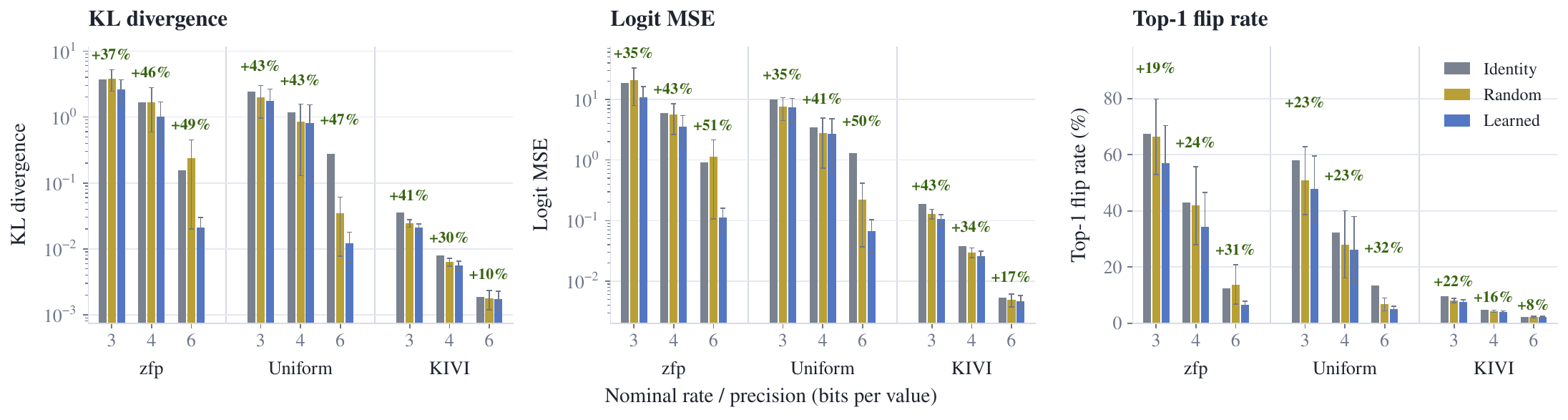}
\caption{Aggregate output-distribution perturbation under $g=16$ gauges at $3$, $4$, and $6$ bits/value. Lower is better; KL divergence and logit MSE use log axes, and top-1 flip rate is a percentage. Error bars summarize the standard error across six model means. Annotations report the mean paired learned-gauge reduction relative to identity coordinates across the six models.}
\label{fig:main-quality}
\end{figure*}

\subsection{Evaluation Design}

We evaluate six checkpoints spanning multiple providers and cache implementations: Qwen3-0.6B, Qwen3.5-0.8B, Llama-3.2-1B, Gemma-3-1B, Phi-4 Mini Instruct, and Ministral-3-3B-Base-2512. The set includes grouped-query softmax-attention decoders, hybrid-cache behavior where only standard tensor KV layers are gauged, and long-context/windowed attention variants. We additionally run a larger Gemma-3-27B model under zfp at the intermediate rates.

Output quality is measured with paired rolling compressed-history scoring over 256 evaluation windows per model. Each window has 2,048 prefix tokens, one teacher-forced rolling input token, and 384 scored target tokens. All conditions consume the same ground-truth input token, but compressed conditions maintain their own histories, so every KV token passes through the condition's gauge, backend, decode, and inverse-gauge path.

Cache-side metrics include compressed bytes, actual bits/value, KV MSE, KV NRMSE, key/value NRMSE, and maximum absolute error. Output-side metrics include paired delta-NLL:
\[
\Delta \mathrm{NLL}
=
\mathrm{NLL}_{\mathrm{compressed}}
-
\mathrm{NLL}_{\mathrm{full}},
\]
as well as $\mathrm{KL}(p_{\mathrm{full}}\|p_{\mathrm{compressed}})$, logit MSE, top-1 flip rate, and top-5 overlap. These paired metrics keep the model, input tokens, and reference logits fixed while measuring loss, distribution shift, and top-$k$ stability. The evaluator writes raw sums and counts. Actual bits/value are computed from measured compressed bytes. Paper figures post-process the released raw files as:
\[
\begin{aligned}
\Delta\mathrm{NLL/token}
&=
\frac{\sum \Delta\mathrm{NLL}}
     {\sum \text{target tokens}},\\
\mathrm{KV\ NRMSE}
&=
\sqrt{
\frac{\sum \mathrm{KV\ SSE}}
     {\sum \mathrm{KV\ reference\ SSE}}
},\\
\mathrm{top1\ flip\ rate}
&=
\frac{\sum \mathrm{top1\ flips}}{\sum \text{target tokens}}.
\end{aligned}
\]
This aggregation is invariant to model-specific batching. Main analyses use 3, 4, and 6 bits/value; 2-bit and 8-bit results remain in the released outputs as boundary conditions.

Sanity rows validate no-compression invariants: identity-clone checks cache cloning, and random/learned inverse rows check the $G^{-1}G$ round trip. In the $g=16$ run, identity clone has zero cache error, while random and learned inverse rows have maximum cache NRMSE below $5.1\times 10^{-8}$.

We also run a serial context-length evaluation for full-cache, zfp-4, block-uniform-4, and KIVI-4 paths. The 4-bit setting provides a matched comparison between fixed-rate numerical compression and low-bit quantization. We record KV allocation growth relative to a one-token origin and total decode-plus-model time normalized by the full-cache path, including compression-path recovery.

For task-prompt checks, we score RULER needle retrieval and variable tracking (\citep{hsieh2024ruler}) together with LongBench-v2 multiple-choice prompts (\citep{bai2025longbenchv2}) under the same zfp rates and coordinate conditions.

\section{Results}

Across implemented backend paths, learned gauges reduce cache reconstruction error and output perturbation. Coordinate controls and held-out spectra identify the source of this effect; a 27B extension and task prompts test its extent, while serial measurements verify the realized storage path.

\begin{figure}[t]
\centering
\includegraphics[width=\columnwidth]{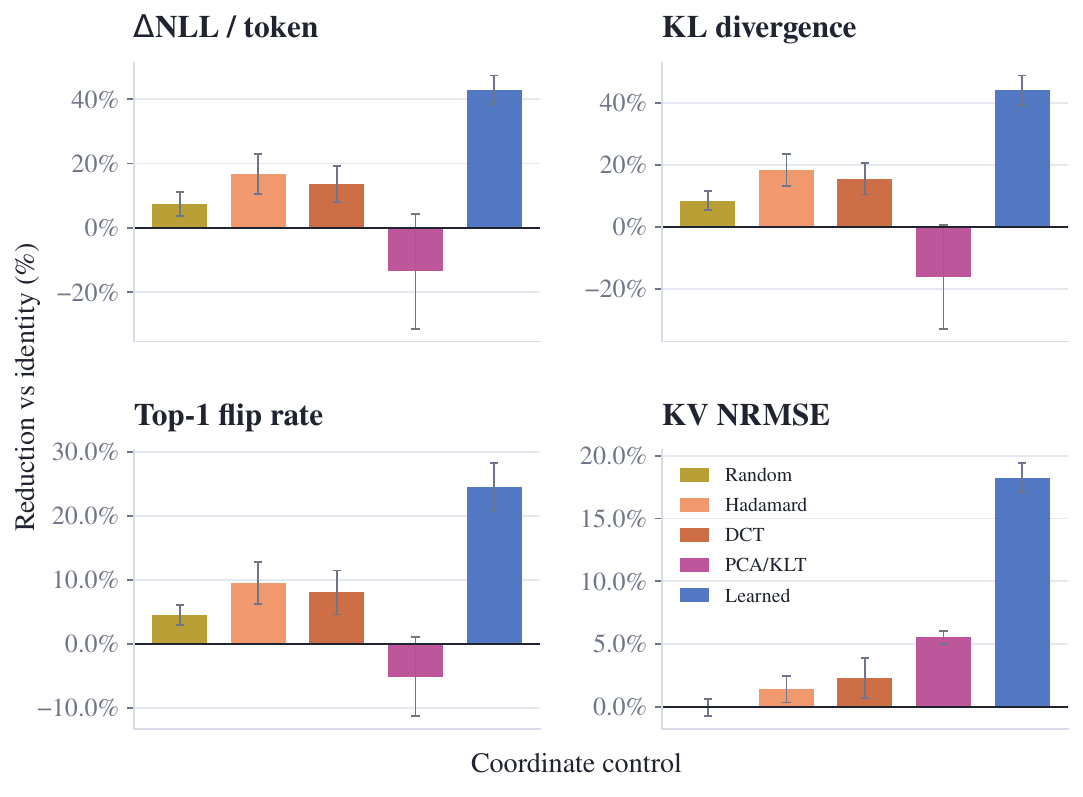}
\caption{Coordinate-control checks for zfp compression at $3$, $4$, and $6$ bits/value. Bars report error reduction relative to identity coordinates; higher is better, and error bars are standard errors across the $18$ model--rate pairs. Random averages three orthogonal gauges. Hadamard, DCT, and PCA/KLT provide fixed or data-derived structured controls.}
\label{fig:coordinate-controls}
\end{figure}

\begin{figure*}[t]
\centering
\includegraphics[width=\textwidth]{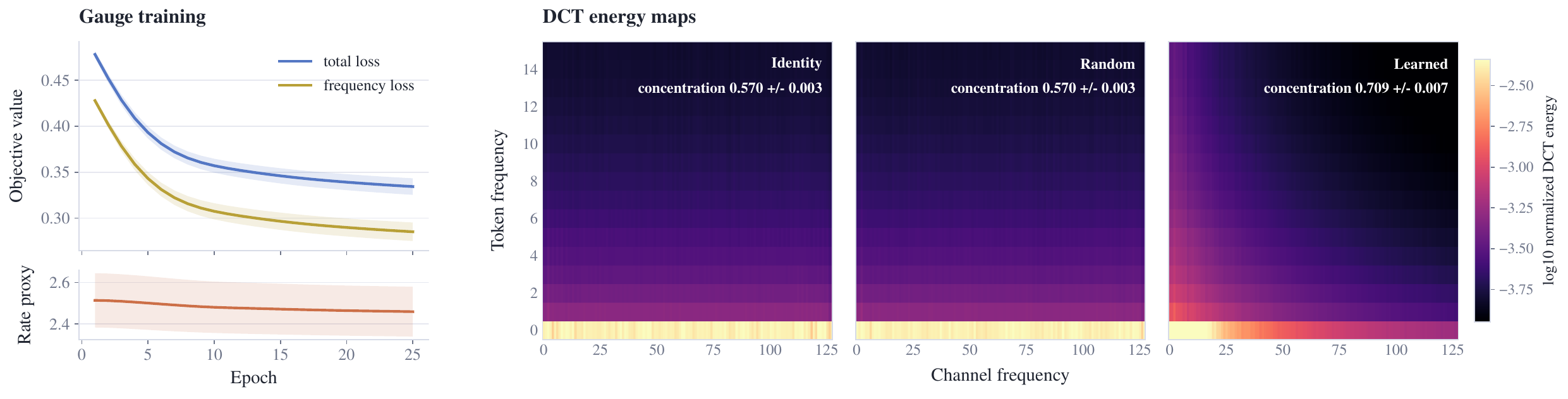}
\caption{Gauge training dynamics and spectral cache geometry. Left: mean objective terms across six models with standard-error bands. Right: representative normalized DCT energy maps from Qwen3-0.6B; the reported concentration means and standard errors aggregate all $268$ standard-attention K/V layer maps across the six models. Learned gauges concentrate energy in lower token--channel frequencies.}
\label{fig:mechanism}
\end{figure*}

\subsection{Compression Quality Under Matched Rates}

The main quantitative analyses use $3$, $4$, and $6$ bits/value. The $2$-bit setting often enters a severe-distortion regime, while $8$ bits/value approaches a high-fidelity boundary where transform-only numerical noise can become comparable to the remaining compression perturbation; both boundary settings are retained in released outputs.

Figure~\ref{fig:main-quality} shows absolute paired quality metrics; percentage labels summarize reductions at matched settings. For fixed-rate zfp compression, learned gauges reduce KL divergence, logit MSE, and top-1 flip rate by $44.0\%$, $43.3\%$, and $24.5\%$ on average relative to identity coordinates. The corresponding zfp actual bits/value are measured from compressed bytes and match the target rates to within $6\times10^{-5}$ bits/value, so the gains come from cache geometry rather than hidden rate differences. Cache reconstruction follows the same trend: learned gauges reduce zfp KV NRMSE by $18.3\%$ over the same operating points.

The same learned gauge also improves scalar low-bit paths. For block-uniform quantization, learned gauges reduce KL divergence, logit MSE, and top-1 flip rate by $44.2\%$, $42.1\%$, and $26.1\%$, respectively. For KIVI-style quantization, the corresponding reductions are $27.2\%$, $31.3\%$, and $15.2\%$. Both quantizers improve all three output metrics in every one of the $18$ model--rate pairs. Thus, a gauge trained only from cache geometry can improve quality preservation for common coordinate-sensitive quantizers without language-modeling, logit, or task-output losses during gauge training.

\subsection{Coordinate Controls}

Figure~\ref{fig:coordinate-controls} evaluates a structured control set under the same zfp backend. Random orthogonal gauges reduce KL by $8.5\%$ on average, Hadamard by $18.3\%$, and DCT by $15.5\%$. PCA/KLT improves KV NRMSE but worsens the output metrics on average, showing that ordering directions by second-moment reconstruction does not determine where decoding error affects attention logits. Learned gauges reduce KL by $44.0\%$ and improve all $18$ model--rate pairs relative to both identity coordinates and the random-gauge average. Among the evaluated controls, the frequency-trained gauge therefore provides the largest and most consistent improvement under matched backend settings and measured rates.

\subsection{Mechanism}

The training objective directly targets frequency concentration in the codec-facing token--channel layout. Figure~\ref{fig:mechanism} confirms that the objective transfers to held-out evaluation windows: identity and random gauges average approximately $0.570$ spectral concentration, while learned gauges reach $0.709$. The total training loss decreases by 30.1\%, and the frequency term by 33.4\%. These measurements connect the training signal to the observed zfp rate-distortion behavior: the gauge creates a more spectrally concentrated storage-facing channel layout for the numerical codec, and actual compression/decompression then yields lower cache and output perturbation.

For scalar quantization, the mechanism also includes range redistribution and outlier smoothing, effects already known to aid low-bit LLM inference (\citep{ashkboos2024quarot,liu2025spinquant,su2025rotatekv}). Learned gauges still reduce output perturbation beyond random gauges in the aggregate quantizer rows, indicating that frequency-shaped geometry contributes useful structure even when the backend's error mechanism differs from zfp's.

\subsection{Scale, Task, and Storage Checks}

\begin{figure}[t]
\centering
\includegraphics[width=\columnwidth]{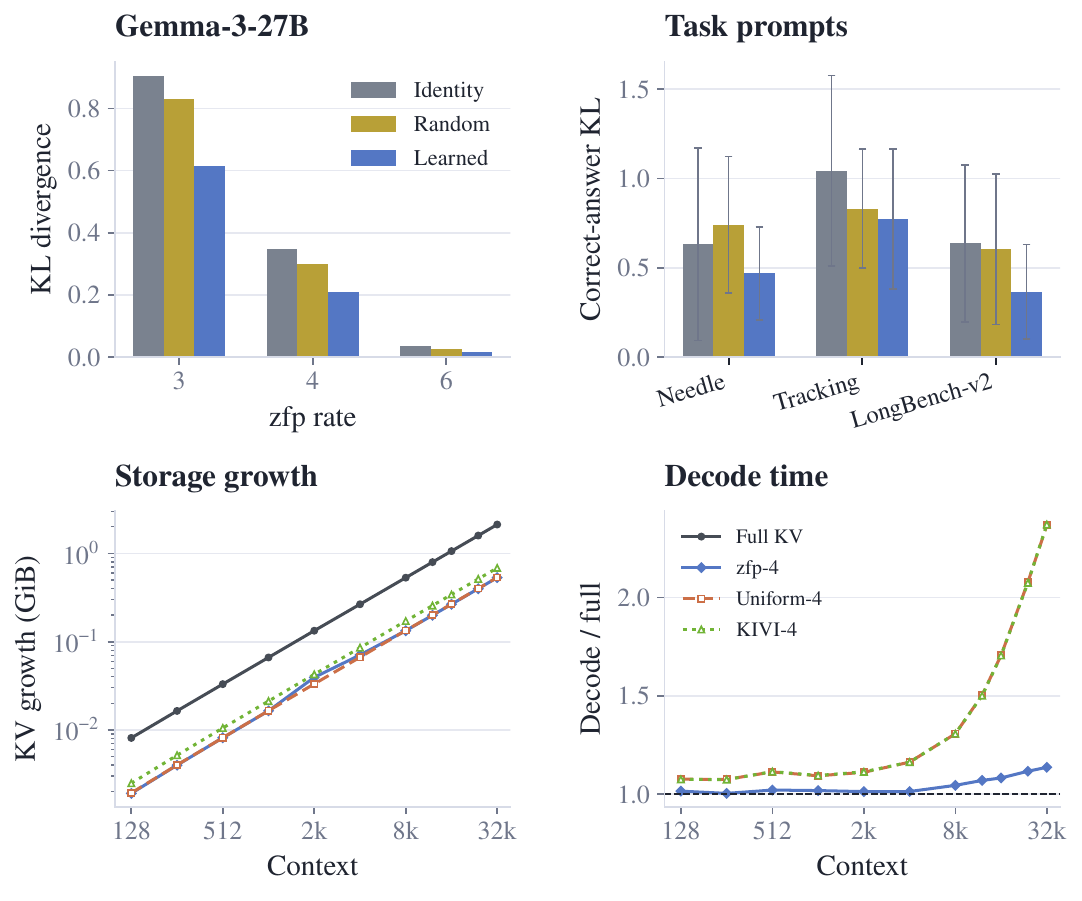}
\caption{Validation beyond the six-model matrix. Top left: zfp KL on Gemma-3-27B. Top right: correct-answer KL averaged over $3$, $4$, and $6$ bits/value for each task family; error bars are standard errors across the three rate-specific model means. Bottom: persistent KV-allocation growth and decode-plus-model time for the learned-coordinate paths in the serial context-length run; timing includes explicit cache recovery.}
\label{fig:validation-system}
\end{figure}

Figure~\ref{fig:validation-system} adds scale, task-prompt, and storage/timing evidence. On Gemma-3-27B, learned gauges reduce zfp KL by $32.1\%$, $39.5\%$, and $49.3\%$ at $3$, $4$, and $6$ bits/value, following the same direction as the six-model matrix.

Task-prompt scoring shows the same paired-distribution trend. Averaged across tasks, models, and rates, learned gauges lower correct-answer KL from $0.772$ to $0.537$ and correct-answer delta-NLL from $0.819$ to $0.571$ relative to the full-cache reference. Accuracy is also recorded, but these moderate-compression settings rarely move the full-cache answer across a decision boundary; likelihood and ranking metrics therefore provide higher-resolution evidence of quality preservation.

The serial context-length run measures persistent KV allocation growth and recovery cost in the implemented path. Allocation is reported as growth over a one-token origin, isolating context-dependent KV memory from fixed model weights and persistent allocator state. At the longest measured context, zfp-4 uses $4.000$ actual bits/value, and its allocation growth is essentially aligned with the 4-bit block-uniform path, while KIVI-4 uses a larger effective payload. Within this serial implementation, zfp-4 has lower measured decode overhead than the quantization paths. These measurements validate that fixed-rate numerical compression attains its intended persistent KV-storage reduction while including explicit cache recovery and model decoding in the measured path.

\section{Discussion and Limitations}

Codec-Gauge improves KV-cache compression quality by training the gauge exposed to a backend. Figure~\ref{fig:coordinate-controls} shows that the improvement is not explained by generic rotation, fixed transform coding, or a data-derived basis alone; Figure~\ref{fig:mechanism} shows the intended storage-facing spectral mechanism. The gauge shapes the codec-facing frequency distribution in a fixed token/channel layout, then is tested through actual compression, decompression, inverse recovery, and rolling-cache scoring.

The quantization rows broaden the finding. Orthogonal rotations and incoherence processing are already strong tools for low-bit quantization, and random gauges capture part of that effect. The learned frequency-distribution objective still reduces output perturbation on top of random gauges for block-uniform and KIVI-style paths, without quantization-specific losses, language-modeling loss, or logit supervision.

The training objective remains outside both codec and model-output loops: it differentiates through neither zfp nor a quantizer and uses no reconstruction or output loss. Its transfer to held-out zfp compression and two scalar quantizers therefore supports a representation-level geometric effect rather than adaptation to one decoder's artifacts.

The practical consequence is fidelity headroom at a fixed cache budget: the same backend configuration and measured rate produce a recovered cache closer to the full-cache reference. Because gauge parameters are fixed per checkpoint and do not scale with context length, their storage cost is amortized as the retained context grows.

The paired rolling metrics measure cache-induced distributional change under fixed tokens and reference logits. KL, logit MSE, delta-NLL, and top-1 flips quantify how much the compressed-cache model deviates from the full-cache model before a discrete answer changes. Prompts from these three task families add likelihood-based checks in long-context task formats; thresholded accuracy is also reported, but is a lower-resolution view when moderate compression rarely changes the selected answer.

The evaluated checkpoints span multiple providers, attention variants, and cache implementations. The 27B model check follows the same trend at a larger scale, while the six-model matrix tests architectural diversity. Results are also stable across $g=8$, $16$, $32$, and full-head gauges; $g=16$ is used as a representative middle setting.

Modern KV-cache systems jointly choose precision, packing, paging, eviction, and scheduling. Codec-Gauge acts at the retained-tensor interface: quantizers and cache managers retain these backend decisions, while the gauge changes only the coordinates presented to them. Holding the remaining system choices fixed therefore isolates the observed rate-distortion improvement as a representation-side effect.

Timing includes explicit key/value recovery rather than assuming fusion. Deployment-oriented implementations can specialize this path through kernel fusion, backend-specific packing, and paged-KV integration without changing the learned coordinate objective. The coordinate controls, two quantization paths, task prompts, 27B extension, and measured storage/timing paths together establish the effect beyond a single model, transform, or evaluation metric.

\section{Conclusion}

We introduced Codec-Gauge, a post-training coordinate-shaping method for Transformer KV caches. Learned gauges improve zfp fixed-rate compression beyond raw and control coordinates under matched measured bits/value, and also improve output preservation for block-uniform and KIVI-style quantization paths. The method keeps the model and backend fixed, learns only small orthogonal KV-coordinate maps from frozen-model cache tensors, and evaluates the result through actual compression, decompression, and rolling-cache scoring.

The broader conclusion is that KV-cache compressibility is not only a property of the backend or the checkpoint in its original basis: the coordinates exposed to a codec or quantizer are an actionable variable. Training this gauge with a frequency-distribution objective improves cache reconstruction and output-distribution stability at storage rates realized by actual compressed-cache paths, without changing attention semantics. Cache-coordinate shaping therefore complements existing KV compression, quantization, and serving pipelines while separating a reusable representation-side optimization from backend-specific engineering choices.

\paragraph{Generative AI use.}
Generative AI tools assisted with manuscript preparation, code, figures, and reference checks; the authors verified all outputs and take full responsibility.

\clearpage
\bibliography{refs}

\clearpage
\appendix
\section*{Appendix}
\setcounter{figure}{0}
\renewcommand{\figurename}{Appendix Figure}
\setcounter{table}{0}
\renewcommand{\tablename}{Appendix Table}

The complete code, configurations, checkpoints, and experiment outputs are available at
\url{https://huggingface.co/cccat6/Codec-Gauge}.
The repository contains raw outputs for the gauge-size sweep, coordinate controls, 27B model check, task scoring, memory diagnostics, reproduction scripts, configurations, checkpoints, generated experiment results, CUDA/zfp build notes, and hardware/software environment records. The released configurations record the exact checkpoint identifiers and experiment settings used for all reported runs.

\begin{center}
\centering
\scriptsize
\resizebox{\columnwidth}{!}{%
\begin{tabular}{@{}ll@{}}
\toprule
Item & Formal run configuration \\
\midrule
GPU & NVIDIA RTX PRO 6000 Blackwell, 97{,}887 MiB VRAM, compute capability 12.0 \\
CPU / memory & AMD Ryzen 9 9950X, 16 cores / 32 threads, 123 GiB RAM \\
OS / driver & Ubuntu 24.04.4 LTS, NVIDIA driver 595.71.05, CUDA 13.2 reported by NVIDIA-SMI \\
Python stack & Python 3.12.3, PyTorch 2.12.1+cu130, Transformers 5.12.1, Datasets 5.0.0 \\
Codec dependency & External CUDA zfp 1.0.1 build linked against CUDA 13 runtime; nvCOMP not used \\
Dataset & FineWeb-Edu sample-10BT train split, text field \texttt{text}, global seed 123 \\
Gauge training & 64 windows, 4,096 tokens/window, 25 gauge epochs \\
Evaluation & 256 windows/model, \texttt{seq\_len}=2,432, \texttt{prefix\_len}=2,049 \\
Gauge sizes & full-head, 32, 16, and 8; main figures use group size 16 \\
Backends & CUDA zfp at 2, 3, 4, 6, 8 bits/value; block-uniform and KIVI-style quantizers at 2, 3, 4, 6, 8 bits \\
\bottomrule
\end{tabular}
}%
\captionof{table}{Formal-run environment and experiment scale used for the released results. The complete sanitized hardware and software record is included in the released artifact.}
\end{center}

\begin{center}
\centering
\scriptsize
\resizebox{\columnwidth}{!}{%
\begin{tabular}{@{}lp{0.74\columnwidth}@{}}
\toprule
Metric & Definition and interpretation \\
\midrule
Actual bits/value & Measured compressed bytes divided by the number of represented KV values; includes codec payloads and quantized representations used by the evaluated path. \\
KV MSE / NRMSE & Reconstruction error between recovered and full-cache KV tensors; NRMSE normalizes the summed squared error by full-cache KV energy. \\
Delta-NLL & Difference between compressed-cache and full-cache teacher-forced negative log-likelihood on the same target tokens. \\
KL divergence & $\mathrm{KL}(p_{\mathrm{full}}\|p_{\mathrm{compressed}})$ between next-token distributions from full-cache and compressed-cache histories. \\
Logit MSE & Mean squared difference between full-cache and compressed-cache logits before softmax. \\
Top-1 flip rate & Fraction of target positions where the highest-probability token under the compressed-cache logits differs from the full-cache highest-probability token. \\
Top-5 overlap & Overlap between the five highest-probability tokens under full-cache and compressed-cache logits. \\
Correct-answer KL / delta-NLL & Task-prompt versions of the distribution and likelihood metrics restricted to the candidate answer or target answer used by the prompt evaluator. \\
\bottomrule
\end{tabular}
}%
\captionof{table}{Metric definitions used in the full-matrix, task-prompt, and storage analyses. The cache metrics measure reconstruction of internal KV tensors, while the output metrics measure how much the recovered-cache model deviates from the full-cache reference distribution.}
\label{tab:metric-definitions}
\end{center}

\begin{center}
\centering
\scriptsize
\resizebox{\columnwidth}{!}{%
\begin{tabular}{@{}lcccc@{}}
\toprule
Backend & KL & Logit MSE & Top-1 flip & KV NRMSE \\
\midrule
zfp & 44.0\% (18/18) & 43.3\% (18/18) & 24.5\% (18/18) & 18.3\% (18/18) \\
Block-uniform & 44.2\% (18/18) & 42.1\% (18/18) & 26.1\% (18/18) & 33.7\% (18/18) \\
KIVI-style & 27.2\% (18/18) & 31.3\% (18/18) & 15.2\% (18/18) & 11.9\% (15/18) \\
\bottomrule
\end{tabular}
}%
\captionof{table}{Exact aggregate learned-gauge reductions for the main $g=16$ matrix. Each cell reports the mean reduction relative to identity coordinates over six models and the $3$, $4$, and $6$ bits/value operating points, followed by the number of model--rate pairs improved out of $18$. The table complements the main figures by making the magnitude and consistency of the effect explicit: zfp and block-uniform improve on every paired condition across all four metrics, while KIVI-style quantization remains positive on average and wins most KV-NRMSE pairs despite its stronger backend-specific quantization structure.}
\end{center}

\begin{table*}[!t]
\centering
\scriptsize
\resizebox{\columnwidth}{!}{%
\begin{tabular}{@{}lcccc@{}}
\toprule
Coordinate control & KL & Logit MSE & Top-1 flip & KV NRMSE \\
\midrule
Random orthogonal & 8.5\% (15/18) & 8.6\% (15/18) & 4.6\% (13/18) & -0.1\% (7/18) \\
Hadamard & 18.3\% (15/18) & 18.4\% (16/18) & 9.5\% (16/18) & 1.4\% (6/18) \\
DCT & 15.5\% (16/18) & 15.1\% (16/18) & 8.1\% (15/18) & 2.3\% (8/18) \\
PCA/KLT & -16.2\% (10/18) & -5.5\% (10/18) & -5.1\% (9/18) & 5.5\% (18/18) \\
Learned gauge & 44.0\% (18/18) & 43.4\% (18/18) & 24.5\% (18/18) & 18.3\% (18/18) \\
\bottomrule
\end{tabular}
}%
\caption{Exact zfp coordinate-control summary at the intermediate operating points. All controls use the same backend, token stream, measured rates, and evaluator; only the cache-coordinate map changes. The fixed and data-derived transforms often help, confirming that cache coordinates matter, but their improvements are uneven across output and reconstruction metrics. The learned gauge is the only control that improves every model--rate pair for all four metrics, supporting the frequency-distribution objective as more than a generic rotation effect.}
\end{table*}

\begin{figure*}[!t]
\centering
\includegraphics[width=\textwidth]{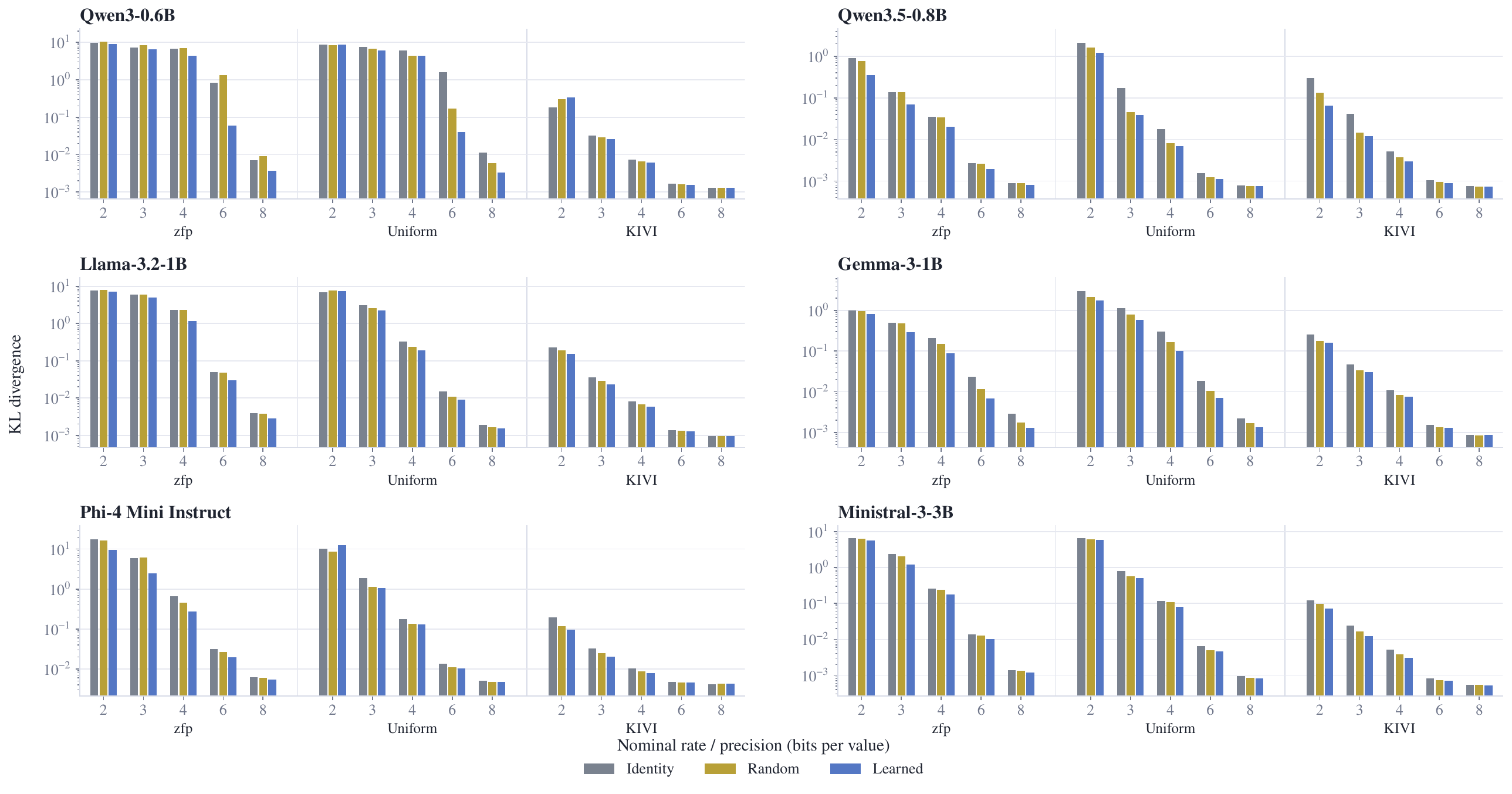}
\caption{Per-model KL divergence for the main $g=16$ matrix across all evaluated zfp, block-uniform, and KIVI settings. Each subplot keeps the model fixed and compares identity, random, and learned coordinates under matched backend settings and measured rates. The learned gauge improves the main trend across model families rather than relying on one checkpoint or one backend setting, while the random-gauge rows show the strength of generic orthogonal controls.}
\end{figure*}

\begin{figure*}[!t]
\centering
\includegraphics[width=\textwidth]{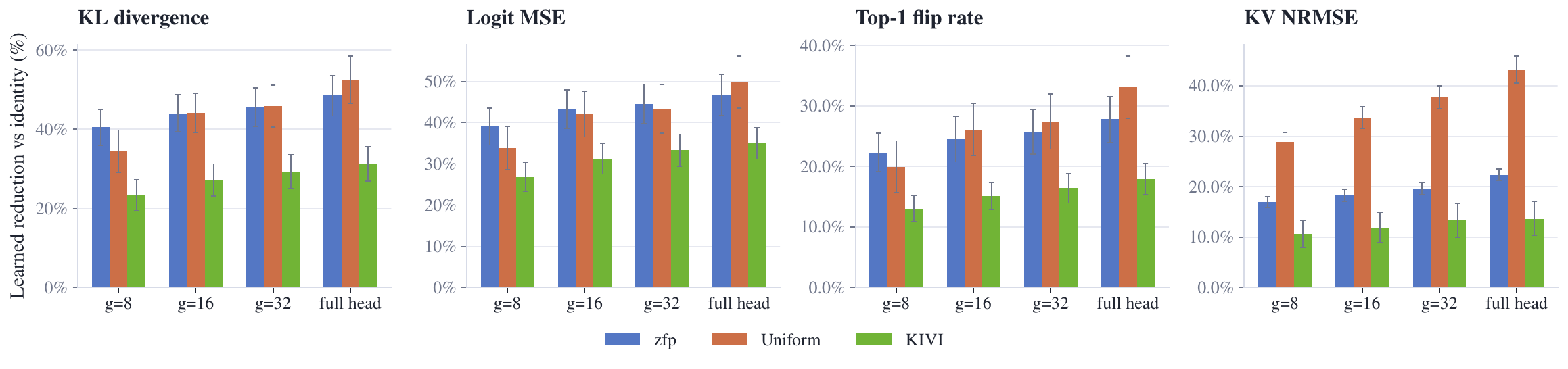}
\caption{Gauge-size sweep over $g=8$, $16$, $32$, and full-head gauges at the intermediate operating points. The same learned-coordinate effect appears across group sizes, showing that the conclusion is not tied to a single matrix dimension. The main text uses $g=16$ as a representative middle setting: it is compact enough to keep the gauge local and inexpensive, while preserving the trend observed in both smaller and larger gauges.}
\end{figure*}

\begin{figure*}[!t]
\centering
\includegraphics[width=\textwidth]{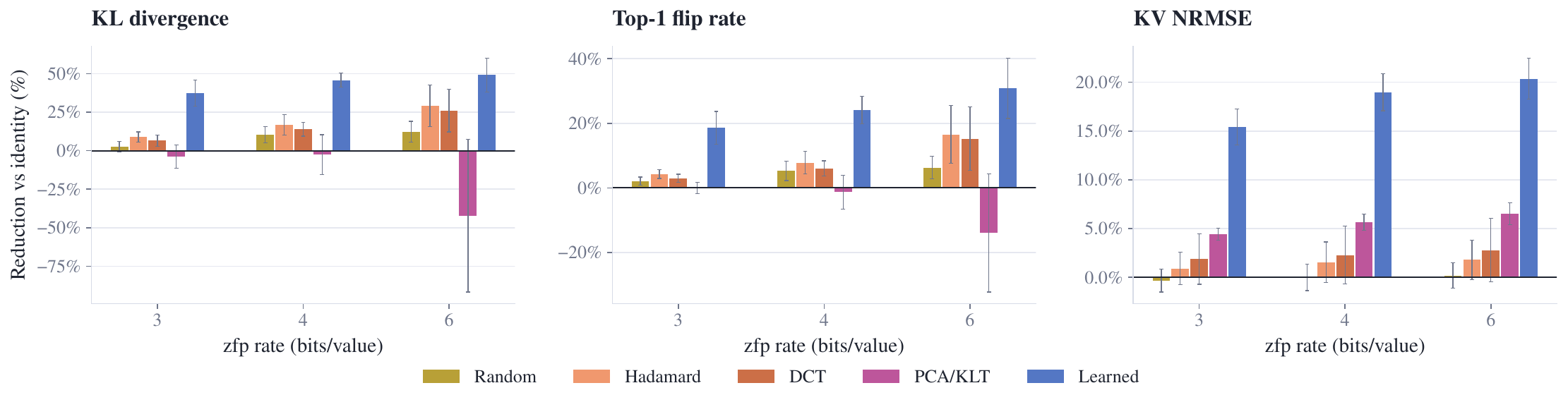}
\caption{Coordinate-control breakdown for zfp by rate, with the identity zero baseline omitted to expose the relative spread among non-identity controls. Random, Hadamard, DCT, PCA/KLT, and learned gauges are all evaluated through the same zfp compression/decompression path. The learned gauge provides the most consistent positive reduction across output perturbation and KV reconstruction metrics, while PCA/KLT illustrates why a reconstruction-oriented data basis alone is not sufficient for the codec-facing objective.}
\end{figure*}

\begin{figure*}[!t]
\centering
\includegraphics[width=\textwidth]{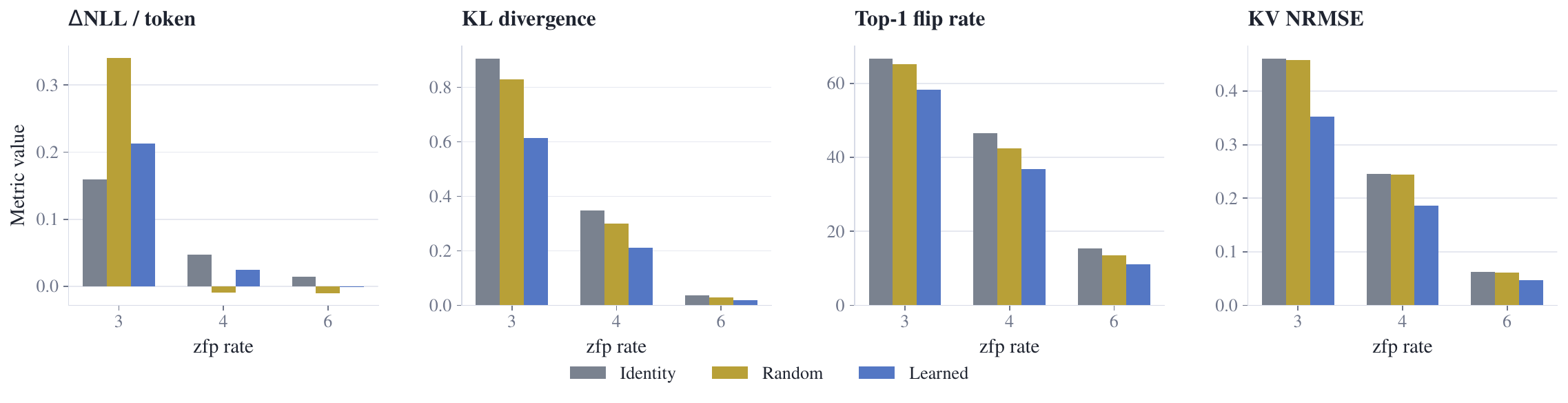}
\caption{Gemma-3-27B metrics under identity, random, and learned gauges. This run uses the same zfp backend and intermediate operating points as the main coordinate study, but on a substantially larger checkpoint. The larger-model check follows the same direction as the six-model matrix, supporting the interpretation that the learned cache-coordinate effect is not restricted to the small-model sweep.}
\end{figure*}

\begin{figure*}[!t]
\centering
\includegraphics[width=\textwidth]{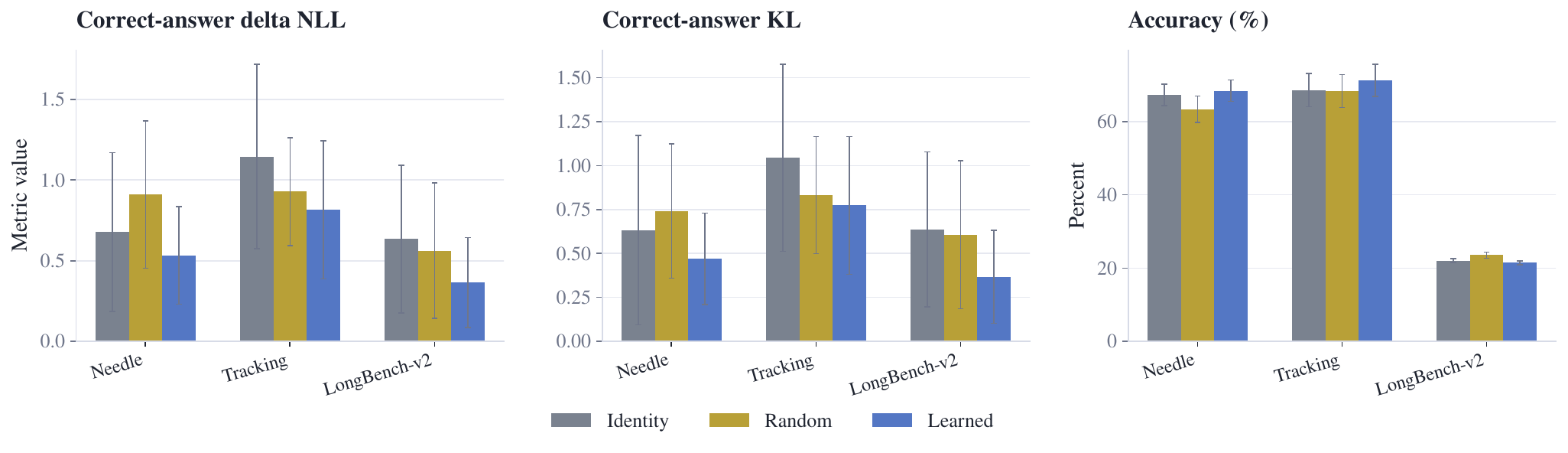}
\caption{Task-prompt scoring breakdown by task family. The prompts cover needle-style retrieval, variable tracking, and LongBench-v2-style multiple-choice checks under the same compressed-cache conditions. Accuracy is included, but the likelihood-based columns are the intended diagnostic signal here: when moderate compression leaves many discrete decisions unchanged, correct-answer likelihood and perturbation metrics reveal the quality margin that accuracy alone can hide.}
\end{figure*}

\begin{figure*}[!t]
\centering
\includegraphics[width=\textwidth]{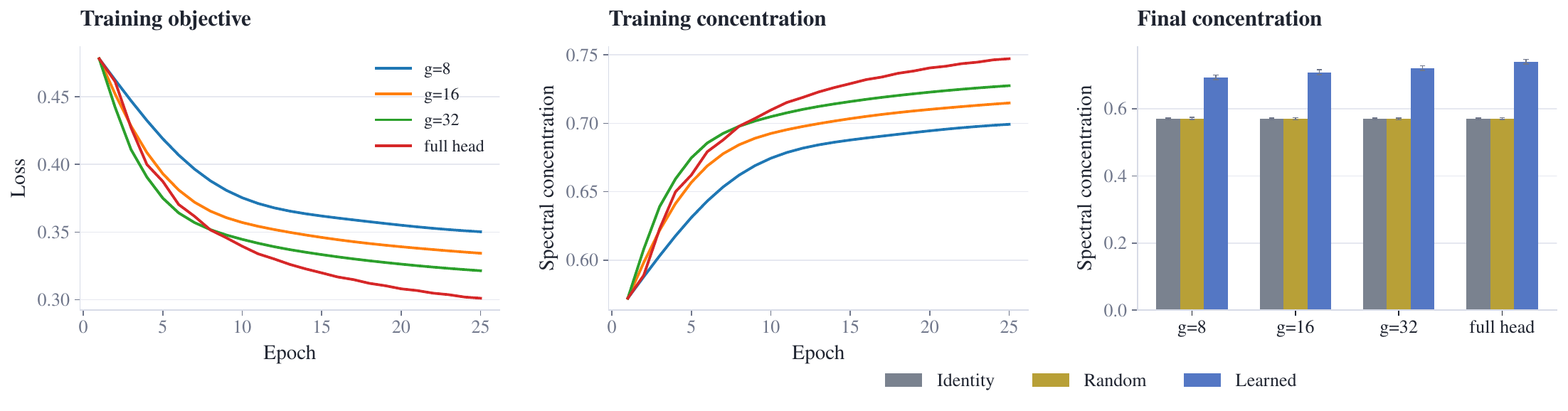}
\caption{Gauge training and spectral concentration across gauge sizes. The loss curves verify that each gauge size learns the intended frequency-distribution objective, and the concentration bars show that the final gauges consistently reshape the codec-facing spectrum. This supports the main mechanism claim: the learned coordinate system changes the numerical geometry seen by the backend rather than only exploiting a single rate or a single model.}
\end{figure*}

\begin{figure*}[!t]
\centering
\includegraphics[width=\textwidth]{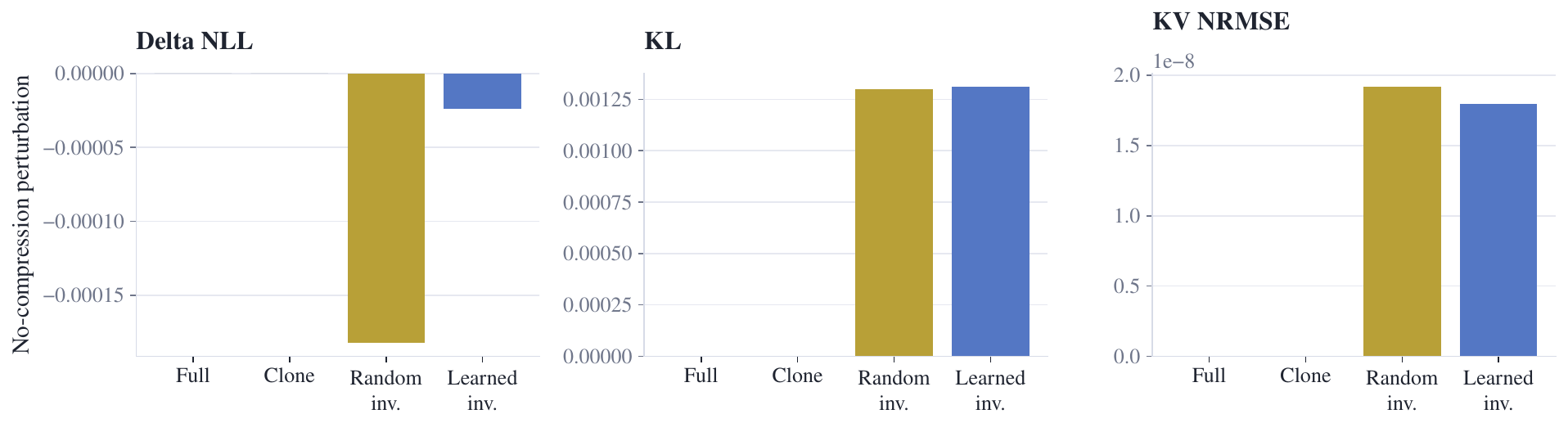}
\caption{No-compression sanity checks for the full cache, cache clone, random-gauge inverse, and learned-gauge inverse paths. These rows isolate implementation and numerical round-trip error without lossy compression or quantization. Perturbations remain near numerical noise, which verifies that the losses reported in the main matrix arise from backend compression or quantization after coordinate shaping, not from the gauge transform itself.}
\end{figure*}

\end{document}